\documentclass{article}
\usepackage{arxiv}
\usepackage[utf8]{inputenc} % allow utf-8 input
\usepackage[T1]{fontenc}    % use 8-bit T1 fonts
\usepackage{hyperref}       % hyperlinks
\usepackage{url}            % simple URL typesetting
\usepackage{booktabs}       % professional-quality tables
\usepackage{amsfonts}       % blackboard math symbols
\usepackage{amsmath}
\usepackage{amssymb}
\usepackage{amsthm}
\usepackage{nicefrac}       % compact symbols for 1/2, etc.
\usepackage{microtype}      % microtypography
\usepackage[square,numbers]{natbib}
\bibpunct[, ]{(}{)}{,}{a}{}{,}
\usepackage{graphicx}
\usepackage{tabularx}
\usepackage{url}
\usepackage[title]{appendix}
\usepackage{algorithm}
\usepackage[noend]{algpseudocode} %does not show end (if, while, for,...)
\makeatletter
\renewcommand{\Function}[2]{%
  \csname ALG@cmd@\ALG@L @Function\endcsname{#1}{#2}%
  \def\currentfunction{#1}%
}
\newcommand{\funclabel}[1]{%
  \@bsphack
  \protected@write\@auxout{}{%
    \string\newlabel{#1}{{\currentfunction}{\thepage}}%
  }%
  \@esphack
}
\makeatother

\RUNAUTHOR{Khodabandeh et al.}

\title{C. H. Robinson Uses Heuristics to Solve Rich Vehicle Routing Problems}

\author{
  Ehsan Khodabandeh\thanks{Corrosponding author} \\
  Opex Analytics\\
  Chicago, IL, USA\\
  \texttt{ehsan.khodabandeh@opexanalytics.com} \\
   \And
 Lawrence V.\ Snyder \\
  Lehigh University\\
  Industrial and Systems Engineering\\
  Bethlehem, PA, USA \\
  \texttt{larry.snyder@lehigh.edu} \\
  \And
 John Dennis \\
  C.H. Robinson\\
  Eden Prairie, MN, USA\\
  \texttt{john.dennis@chrobinson.com} \\
  \And
 Joshua Hammond \\
  C.H. Robinson\\
  Eden Prairie, MN, USA\\
  \texttt{joshua.hammond@chrobinson.com} \\
  \And
 Cody Wanless \\
  C.H. Robinson\\
  Eden Prairie, MN, USA\\
  \texttt{cody.wanless@chrobinson.com} \\
}

\begin{document}
\maketitle

\begin{abstract}
We consider a wide family of vehicle routing problem variants with many complex and practical constraints, known as rich vehicle routing problems, which are faced on a daily basis by C.H. Robinson (CHR). Since CHR has many customers, each with distinct requirements, various routing problems with different objectives and constraints should be solved. We propose a set partitioning framework with a number of route generation algorithms, which have shown to be effective in solving a variety of different problems. The proposed algorithms have outperformed the existing technologies at CHR on 10 benchmark instances and since, have been embedded into the company's transportation planning and execution technology platform. 
\end{abstract}


\section{Introduction}
C.H. Robinson (CHR) is a Fortune 500 company that solves logistics problems for companies across the globe and across industries, from the simple to the most complex. With over $\$20$ billion in freight under management and 18 million shipments annually \citep{CHR}, C.H. Robinson is the world’s largest logistics platform. CHR's global suite of services accelerates trade to seamlessly deliver the products and goods that drive the world’s economy. With the combination of their multi-modal transportation management system and expertise, they use their information advantage to deliver smarter solutions for more than 124,000 customers and 76,000 contract carriers.

CHR is one of the world's largest third-party logistics (3PL) providers. A third-party logistics organization provides value to its customers by arranging for transportation of freight commodities from a shipper/consignor (the \textit{first party}) to a receiver/consignee (the \textit{second party}). The 3PL company coordinates the pickup and delivery of the commodities, though they may not own the equipment used for transportation. 
3PLs often provide additional services, including freight consolidation, replenishment, customs brokerage, supply chain network analysis and design, reverse logistics programs, and business process consulting/outsourcing.

CHR's customers include shippers across many industries, including consumer packaged goods (CPG), food and beverage, retail, manufacturing, chemicals, automotive, paper, electronics, and more. 
CHR is considered an ``asset-light'' 3PL, which means that they don't own any transportation equipment, but instead broker transportation using carriers from their partner network to their customers' freight.

In organization of such magnitude, reducing cost is essential. To do so, CHR carefully pairs and ships customer orders. Such pairing requires that many constraints be satisfied: orders' specified pickup and delivery time windows cannot be violated; total orders on a route cannot exceed their truck's capacity; working and driving hours regulations must be respected; the total number of stops per truck should be less than a maximum number; and many more that depend on the customer, mode of transport, and carriers.

To combine orders and find the most efficient routing solution, C.H. Robinson load planners leverage Navisphere\textsuperscript{\textregistered}, CHR's technology platform. This platform facilitates simple load consolidation, aggregating orders based on certain specified criteria.
In addition to the capabilities within the Navisphere\textsuperscript{\textregistered} platform, CHR also leverages several commercially available vehicle routing software packages. There are several drawbacks to the existing capabilities and process. The off-the-shelf packages do not natively integrate with Navisphere\textsuperscript{\textregistered}, require vendor-specific expertise, and often cannot handle complex constraints. Additionally, the process may require manual adjustments, and frequently relies too much on the planner's experience.

The main problem with which CHR is concerned is a variant of the Vehicle Routing Problem (VRP), known as the Multi-Attribute Vehicle Routing Problem (MAVRP) or the Rich Vehicle Routing Problem (RVRP). RVRPs typically combine many complex constraints designed to help tackle realistic problems. Due to the nature of CHR's business, each customer's specific RVRP can be very different from the others (e.g., one customer may impose a limit on the number of stops of each truck, another may forbid visiting a certain sequence of locations, etc.).
Throughout this project, CHR planners, analysts, and software developers have worked with analysts at Opex Analytics, with guidance from a faculty member at Lehigh University, to develop a customized routing tool that would solve many types of RVRP instances with different sizes and constraints.
This paper will not demonstrate that we have successfully outperformed the best standard results on RVRPs in the literature, but instead will explain algorithms that perform well on a variety of real-world RVRPs, as well as how we have embedded them into a customized tool for CHR.

\citet{lahyani2015rich} presented a taxonomy and definition of RVRPs and introduced a new classification scheme. \citet{caceres2015rich} surveyed the latest advances in the field and summarized problem combinations, constraints, and approaches. There is also a comprehensive survey of heuristics for MAVRPs up to 2013 in \citet{vidal2013heuristics}. Later, \citet{vidal2014unified} introduced a Unified Hybrid Genetic Search (UHGS) metaheuristic for solving different variants of MAVRPs. Their algorithm relies on problem-independent unified local search, genetic operators, and advanced diversity management methods to increase the effectiveness of the local search. The authors conducted extensive experiments and reported that in 1045 of the 1099 best-known solutions, UHGS matched or outperformed the state-of-the-art problem-tailored algorithms.

Many researchers have also focused on solving real-world RVRPs. \citet{pellegrini2007multiple} used a framework called Multiple Ant Colony Optimization to solve an RVRP for an Italian firm that delivers a wide number of food products to restaurants and retailers in northeast Italy. The problem had multiple time windows, a heterogeneous fleet, a maximum duration for subtours, multiple visits, and multiple objectives.

\citet{amorim2014rich} solved a heterogeneous fleet site-dependent VRP with multiple time windows for a Portuguese food distribution company using the adaptive large neighborhood search framework. By achieving better capacity utilization for the company's vehicles as well as reducing the total distance traveled to customers, they demonstrated that the company's cost could be reduced by 17\% during peak seasons.

\citet{lahyani2015multi} introduced, modeled, and solved a rich multi-product, multi-period, and multi-compartment VRP with a required compartment cleaning activity. They proposed an exact branch-and-cut algorithm to solve the problem. The authors evaluated the performance of the algorithm on real-life data sets of olive oil collection processes in Tunisia, under different transportation scenarios. Instances with one depot and up to 45 transportation requests loaded in three or four vehicles could be solved to optimality.

\citet{de2015gvns} presented a general variable neighborhood search (VNS) metaheuristic for solving a VRP with a fixed heterogeneous fleet of vehicles, soft and multiple time windows, customer priorities, vehicle-customer constraints, and several objective functions. The proposed solution has been embedded into the fleet management system of a company in the Canary Islands. Later, \citet{de2015variable} also considered a dynamic RVRP where customers' requests can be either known at the beginning of the planning horizon or dynamically revealed over the day.

\citet{sicilia2016optimization} proposed an algorithm based on VNS and Tabu Search (TS) to solve the problem of goods distribution, a problem originally faced by a large Spanish distribution company operating in major urban areas throughout Spain. The company had to consider capacity, time windows, compatibility between orders and vehicles, a maximum number of orders per vehicle, and site-dependent pickup and delivery constraints. The main objective was reducing costs caused by inefficiency and ineffectiveness. The proposed algorithm has been integrated into a commercial software tool, which is used daily.

\citet{osaba2017discrete} studied the problem of a medium-sized newspaper distribution company in Bizkaia, Spain. The company in question faces a multitude of constraints, including a strict recycling policy, a requirement to treat each town/city separately, avoiding certain forbidden streets, and accommodating variable travel times. The authors developed a discrete firefly algorithm to solve this RVRP. They compared their approach with both an evolutionary algorithm and an evolutionary simulated annealing approach, each using a benchmark of 15 instances with 50 to 100 customers, and showed promising results.

The remainder of the paper is organized as follows. In the next section, we describe the problem formulation and its assumptions. After that, we present our solution methodology and the motivations behind it. We then discuss the performance of various methods on CHR's test datasets. Finally, we discuss conclusions and future work.

\section{Problem Statement}
CHR receives requests from its customers for handling their orders. Typically, the pickup and delivery locations of an order, the specific time windows during which the order can be picked up and dropped off, and the order's size specifications (e.g., weight, volume, and number of pallets required in a truck) are specified in the request. The problem CHR faces is how to best pair and route different orders such that their total transportation cost and number of required trucks are minimized. In addition, the trucks do not need to return to their origin location (i.e., the return is not considered as part of the transportation cost). This is a classic \textit{open} VRP. Moreover, there are other constraints that, depending on the specific problem at hand, should be respected:
\begin{itemize}
    \item There are usually multiple carriers that can handle CHR's orders, each with its own fleets and specifications. For example, some carriers may not serve certain regions (e.g., specific cities or states), or may have a limited fleet of trucks with a certain capacity. In this case, we are dealing with \textit{heterogeneous fleets}. 
    \item There are many types of transportation services with different capacity constraints. We dealt with a number of them, including the three most common: truckload (TL), less-than-truckload (LTL), and intermodal. Consequently, we refer to this as a  \textit{multi-modal} network. 
    \item \textit{Driver regulations} are imposed by the Department of Transportation's (DOT) hours-of-service rules \citep{DOT}. For example, drivers are not permitted to drive for more than 11 hours or work for more than 14 hours in a given day, and must have at least 10 off-duty hours after a full day of work.
    \item There are some \textit{order, site, customer, and vehicle-dependent} constraints, such as:
    \begin{itemize}
        \item \textit{Incompatibility constraints} may occur between orders and vehicles. For example, distribution of different types of food may require vehicles with different temperature levels. Also, some products cannot be transported together in the same vehicle.
        \item Customers may require that certain orders or locations are visited first or last on a route.
        \item Customers may impose \textit{regional (in)compatibility constraints} (e.g., orders from certain cities or states should (not) be paired with each other).
        \item There is a limit on the number of pickups or deliveries a truck can make.
        \item There are also \textit{distance-related} constraints, including caps on maximum total distance, maximum out-of-route (OOR) distance, maximum out-of-route percentage, maximum distance between the first and the last pickups, or maximum distance between the first and last drops.
    \end{itemize}
\end{itemize}

In this paper, we use the terms \textit{route} and \textit{truck} interchangeably. Two of the terms above may need further explanation: OOR distance and OOR percentage. In multi-stop routes, the OOR distance is the difference between the total distance and the direct distance from the origin to the final destination. The OOR percentage is the OOR distance expressed as a percentage of the direct distance.

As some of CHR's customers are interested in minimizing total cost, some in minimizing total distance, and some in maximizing truck utilization, any suggested methodology should be flexible enough to work on each of these instances of RVRPs. In the next section we focus on the overall methodology to solve CHR's RVRP instances.

\section{Solution Methodology}
Since many of CHR's routing problems are heavily constrained, our solution incorporated the set partitioning (SP) formulation of VRP, originally introduced by \citet{balinski1964integer}. For a description of the formulation, see the set partitioning model \ref{sec:sp_model} in the appendices.

Not only does the SP model allow for both general and flexible cost structures (as well as any other non-cost-based objective functions), but it also provides an easy way to add side constraints (e.g., limiting the total number of routes, or capping the number of trucks of certain type or capacity). Moreover, we do not need to concern ourselves with the feasibility of a route in the SP model when new constraints are imposed, because the route validation happens prior to solving the SP model. The process starts by generating a set $J$ of routes and then solving the SP model using that set, and the route generation process includes all of the logic for validating the feasibility of a route.

However, one main drawback of using the SP model for VRP is its very large number of variables, especially in non-tightly-constrained (NTC) instances \citep{toth2002vehicle}. Generating all the feasible routes in NTCs or other large problems may not be possible, so the primary challenge is to generate good routes. Henceforth, we use ``hard''/``harder'' to describe NTC or larger problem instances and ``easy''/``easier'' to describe more easily solvable problems. Next, we discuss how to develop route generation algorithms for CHR's RVRPs.

\subsection{Exact Method}
Because our initial problem instances in the project had only a few hundred orders and were tightly constrained, we started with an exhaustive search algorithm that generated all feasible multi-stop routes (hence the name ``Exact Method'') and then solved an SP model. The advantages of this approach were threefold:
\begin{enumerate}
    \item We could solve instances with $100+$ and even a few instances with $400+$ orders to optimality. CHR does, in fact, have customers with only a few hundred orders, and this approach can be useful for such small customers.  In most cases we could achieve that in a few seconds and no more than a minute.
    \item In harder instances where this method's run time was not acceptable, we could still allow much larger run time limits and use the obtained solution as the baseline for judging the quality of the heuristic methods' solutions.
    \item When generating all the routes was impossible, we could investigate the constraints and focus on the bottlenecks to develop more fine-tuned heuristic algorithms.
\end{enumerate}

Note that we use multi-stop routes to refer to either routes with one pickup and multiple drop-off locations (1PMD) or routes with multiple pickup and one drop-off locations (MP1D). The way our algorithms are set up makes creating MP1D routes similar to creating 1PMD routes. For a 1PMD route, we start from a pickup location and then consider visiting the drop-off locations that have an order originating from that pickup location. This continues until we reach the limit on the number of drops, assuming all the other constraints are valid. In a MP1D route, the process is mirrored: we start from a drop-off location that has orders originating from different pickup locations. The pickup locations are visited until we reach the limit on the number of pickups, assuming the validity of all the other constraints. So, without loss of generality, we base all our explanations on 1PMD routes.

To illustrate the overall approach for generating all feasible multi-drop routes, consider a small example depicted in Figure~\ref{Figure1}. For a more formal description of this method, see the pseudocode in Algorithm~\ref{alg:md} in the appendices.

\begin{figure}
    \centering
    \includegraphics[width=0.7\linewidth]{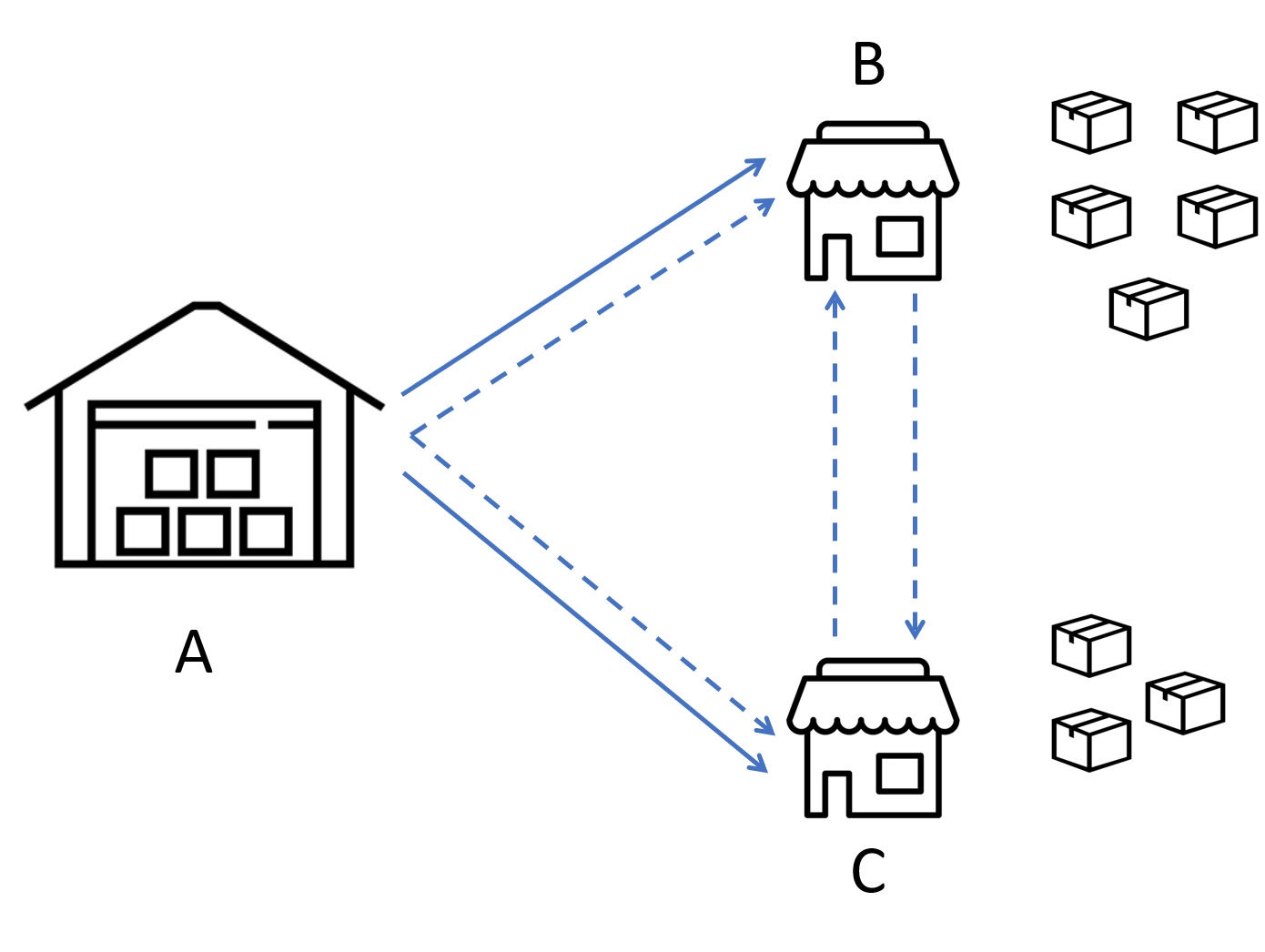}
    \caption{Small network with one pick location (A) and two drop locations (B and C). Solid lines show direct shipment and dashed lines show the direction of multi-drop routes.}
    \label{Figure1}
\end{figure}

As seen in Figure~\ref{Figure1}, five orders should be shipped from location A to B and three orders need to go from A to C. To generate all the routes that ship the orders from A to B, we generate all five routes that only have one order, all 10 routes with a pair of orders, \ldots, and one route with all five orders (Function~\textproc{1P1D Routes} in Algorithm~\ref{alg:md}). Although we are enumerating all the possible routes, in this step we consider the possibility of consolidating the orders that are going from \textit{one} pick location to \textit{one} drop location. Therefore, we call this step \textit{order consolidation}.
Essentially, we create a total of $2^L-1=2^5-1=31$ routes, where $L$ is the number of orders going from one origin to one destination. Similarly, $2^3-1=7$ routes are generated for orders going from A to C. All these routes are validated, and are only accepted if feasible.

Since there are only two destinations (B and C), each route can have either location as its first or second drop. Assuming both possibilities are feasible, we need to combine orders of different locations and create all the routes (Function~\textproc{1PMD Routes} in Algorithm~\ref{alg:md} takes care of that). New routes with $M$ drops are generated by adding new locations at the end of the feasible $(M-1)$-drop routes. For A-B-C routes, we consider each of the seven A-C combinations as a potential addition to each of the 31 A-B routes (this process is similar for A-C-B routes as well). Thus if no constraint is violated, $31 \times 7=217$ A-B-C and 217 A-C-B routes are created. 

With $n$ distinct destinations and $K$ drops allowed, we inherently consider all the $k$-permutations of $n$ destinations (${P_k}^n$), with $k \in 2,\ldots,K$. Since this step deals with exploring and adding orders of different locations, we call it the \textit{neighborhood search} step. Ultimately, all the 1P1D and 1PMD routes are passed to a set partitioning model.

Both order consolidation and neighborhood search are time-consuming steps of the process, since we consider every \textit{combination} of orders and every \textit{permutation} of locations to visit. Knowing these bottlenecks can help us develop alternative route generation approaches in the hope of producing \textit{good routes}.

\subsection{Heuristic Algorithms}
We categorize these heuristics into two classes: a) order consolidation heuristics and b) neighborhood search heuristics. 

\subsubsection{Order Consolidation Heuristics.}
In an order consolidation problem, we try to fill up trucks with orders that are all going from the same origin to the same destination. This  reduces to a one-dimensional bin packing problem (BPP), which many algorithms can solve exactly \citep{delorme2016bin} and approximately \citep{coffman2013bin}. Our goal is not to find the \textit{best} solution to the BPP, but rather to generate \textit{many} good solutions to serve as routes in the SP problem (either as individual routes or for generating multi-drop routes). Consequently, the exact solution to our order consolidation problem might overlook many good multi-drop solutions. Having solved two BPP models for 1P1D routes in the above mentioned example, we may only need one route from A to B and one from A to C to cover all the orders. This means there are only two possible multi-drop options, and therefore a total of four possible routes. Compare this with $31 + 7 = 38$ 1P1D and $2\times(31 \times 7) = 434$ 1PMD options in the exhaustive enumeration.

To generate diverse and sufficient order consolidation options yet avoid complete enumeration, we solve the BPP using a few simple heuristics:
\begin{itemize}
    \item \textbf{First Fit Decreasing (FFD):} The items are first sorted in decreasing order of their sizes, and then each item is assigned to the lowest-indexed truck with sufficient space, or to a new truck if it doesn't fit in an existing one \citep{johnson1973near}
    \item \textbf{Best Fit Decreasing (BFD):} Similar to FFD, but an item is placed in a truck where it leaves the smallest remaining space, or in a new one if it doesn't fit in an existing truck \citep{johnson1973near}
    \item \textbf{First Fit Shuffled (FFS):} This is the First Fit algorithm, except the items are first shuffled
    \item \textbf{Singletons:} Each item is assigned to a new truck by itself
\end{itemize}

Although not a method, we regularly use a parameter called \textit{Partial Container} in each of these methods. This parameter is between $(0,1)$, and scales a truck's size to reserve capacity in multi-stop routes. For example, we can multiply a truck's size by $0.5$ and solve with the BFD algorithm.
One must note that although FFD, BFD, and FFS can all be used separately or together to solve the order consolidation problem, the \textit{Singletons} method is \textit{never} used by itself. As the most naive approach, its only purpose is to diversify the solution pool and increase the chances of creating more (and hopefully better) multi-stop routes later.

With the same idea in mind, one can also combine these heuristics with the Exact Method and generate order consolidation options conditionally. In other words, we can define a threshold parameter ahead of time, and if the number of orders going from an origin to a destination is \textit{less} than that threshold, we generate the order consolidations using the Exact Method. If not, we can use one or more 1P1D heuristic approaches.

\subsubsection{Neighborhood Search Heuristics.}
We propose two heuristics to substitute the complete neighborhood search of multi-drop route generation: one that we call ``K-Nearest Neighbors'' (KNN), which is a generalization of the well-known nearest neighbor search, and one that we call ``K-Closest On-the-Route Neighbors'' (K-CORN).

\noindent\textbf{K-Nearest Neighbors (KNN)}

For every location A, sort all other locations in ascending order of their distances from A. By selecting the first $K$ locations, we obtain the $K$ neighbors nearest to A. Therefore, rather than evaluating every destination as the new last stop of a given route, we only consider the $K$ locations that are closest to the current final destination.

\noindent\textbf{K-Closest On-the-Route Neighbors (K-CORN)}

The only difference between K-CORN and KNN is how the neighbors are constructed. In KNN, the distances between locations are used to form the neighborhood (in theory, if two locations are relatively close to each other, they should be neighbors). 
In K-CORN, however, the neighbors are defined using OOR distance. Adding a new stop at the end of an existing route is only considered if this new addition imposes a sufficiently small OOR distance. A location with a small OOR distance from an existing stop on the route is considered to be essentially \textit{on the route}, as the truck only needs to deviate slightly from its path to include it. 

With this neighborhood formation criterion in place, for every location A, sort all other locations by their OOR distances from A in ascending order, and then select the first $K$ locations to obtain the $K$ closest on-the-route neighbors of A.

In all these algorithms, the order of constraint validation and attempts to reduce computational redundancies are big components in generating good routes. For example, if route A-B-C is infeasible due to violation of total distance, then A-B-C-D will be infeasible as well. As problem instances become more complex, both route generation and validation can become computationally expensive. This algorithmic engineering is a crucial part of obtaining high-quality solutions.

\section{Computational Results}
These algorithms were first tested on several pilot datasets provided by CHR, and their results were compared with the Exact Method.

All the algorithms were coded in Python 3.6.8 (single thread), and the set partitioning model is solved using CPLEX 12.9. All the computational tests in this section were run on a $2.80\,\mathrm{GHz} $ Intel Core i7 laptop with 16 GB of RAM.

\begin{figure}
    \centering
    \includegraphics[width=1.0\linewidth]{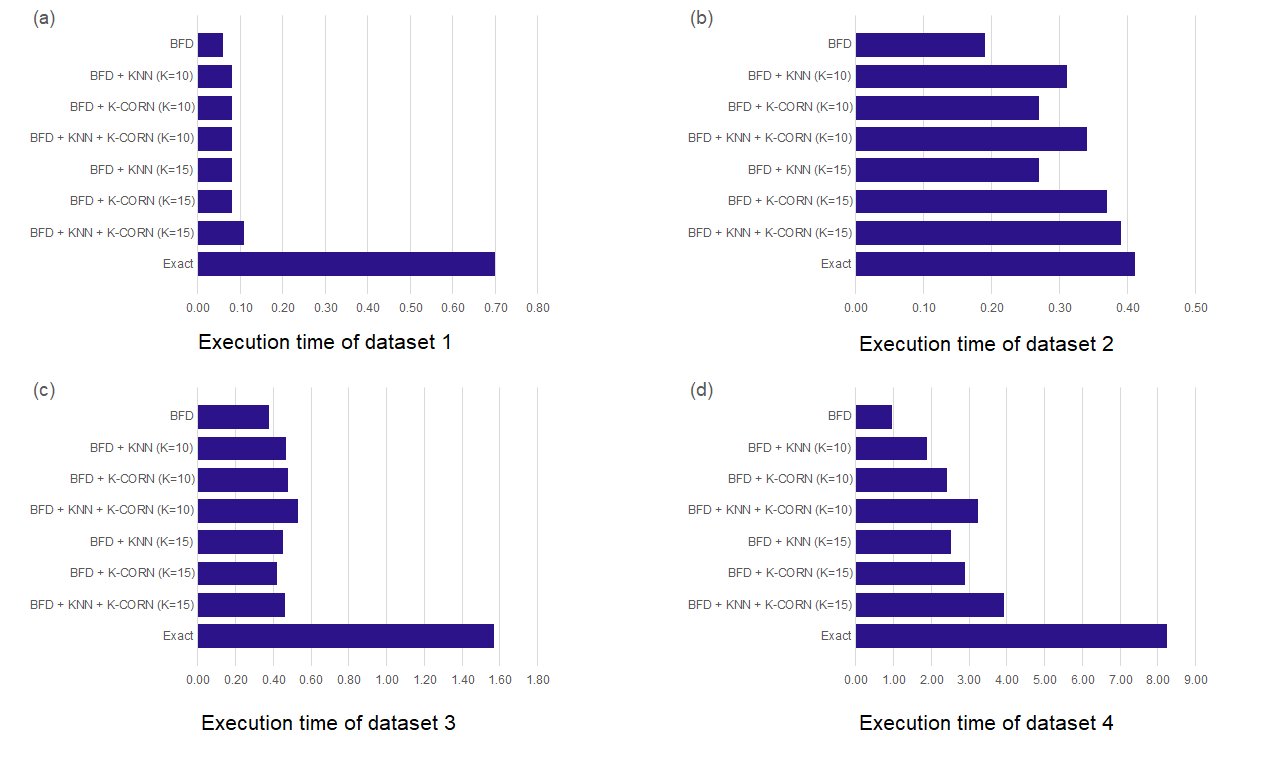}
    \caption{Execution time (in seconds) of different algorithms on the pilot datasets}
    \label{Figure2}
\end{figure}

\begin{figure}
    \centering
    \includegraphics[width=1.0\linewidth]{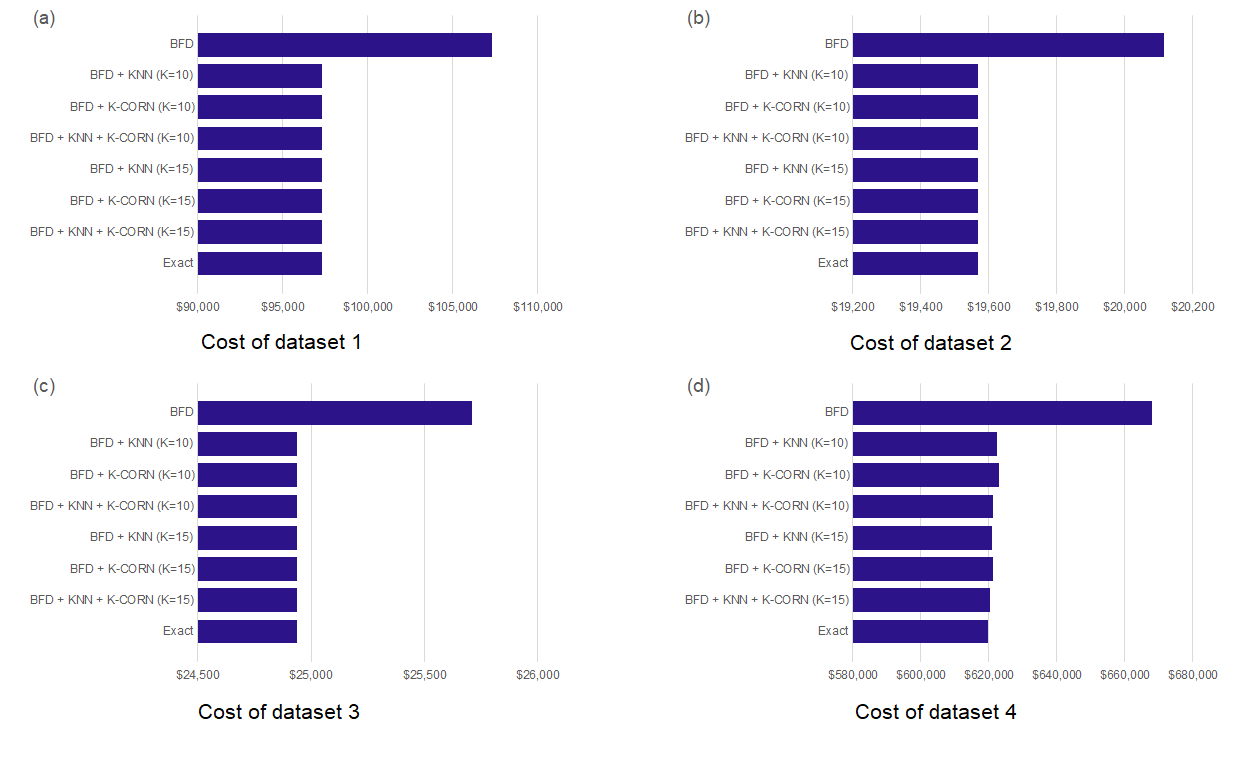}
    \caption{Objective value (cost) of different algorithms on the pilot datasets}
    \label{Figure3}
\end{figure}

The performance of an algorithm is deemed acceptable if its objective function value on the pilot datasets is within $5\%$ of the Exact Method's value.
Table \ref{tab:pilot-features} shows a summary of important features of each dataset, and Table \ref{tab:pilot-results} shows the percentage \textit{relative gap} of each algorithm compared to the Exact Method. Figure~\ref{Figure2} and Figure~\ref{Figure3} show the execution times and costs, respectively.

\begin{table}
    % \small  % if we use this, the whole table will have small font
	\centering
	\caption{Features of pilot datasets}
	\label{tab:pilot-features}
    \begin{tabular}{l|l|l|l|l}
         & \multicolumn{4}{c}{\textbf{Datasets}} \\ \cline{2-5} 
        \textbf{Features} & \textbf{1} & \textbf{2} & \textbf{3} & \textbf{4} \\ \hline
        No. of Orders & 73 & 113 & 184 & 737 \\
        No. of Transportation Modes & 2 & 2 & 5 & 5 \\
        Min Order Weight & 157 & 153 & 4.4 & 1,585 \\
        Avg Order Weight & 9,150 & 2,038 & 1,868 & 38,912 \\
        Max Order Weight & 19,788 & 28,120 & 21,677 & 47,000 \\
        Total Order Weight & 667,987 & 230,348 & 343,732 & 28,678,351 \\
        No. of Origins & 2 & 2 & 3 & 3 \\
        No. of Destinations & 21 & 104 & 101 & 273 \\
        Smallest Truck Capacity & 20,000 & 20,000 & 15,000 & 12,000 \\
        Largest Truck Capacity & 20,000 & 42,000 & 39,000 & 47,000 \\
        Max No. of Drops* & 4 & 2 & 2 & 4 \\
        Max Distance (mile)* & NA & NA & NA & NA \\
        Max OOR Distance (mile)* & 400 & 400 & 200 & 500 \\
        Max First to Last Drop Distance (mile)* & NA & 125 & 1,000 & NA \\
        Avg Delivery Time Window Span (day) & 4.67 & 9.89 & 1.34 & 3.87
    \end{tabular}
\end{table}

\begin{table}
	\centering
	\caption{Percentage relative gap of each algorithm compared to the Exact Method for the pilot datasets}
	\label{tab:pilot-results}
    \begin{tabular}{l|l|l|l|l}
         & \multicolumn{4}{c}{\textbf{Datasets}} \\ \cline{2-5} 
        \textbf{Algorithms} & \textbf{1} & \textbf{2} & \textbf{3} & \textbf{4} \\ \hline
        BFD & 10.31\% & 2.80\% & 3.11\% & 7.75\% \\
        BFD + KNN (K=10) & 0.00\% & 0.00\% & 0.00\% & 0.41\% \\
        BFD + K-CORN (K=10) & 0.00\% & 0.00\% & 0.00\% & 0.52\% \\
        BFD + KNN + K-CORN (K=10) & 0.00\% & 0.00\% & 0.00\% & 0.20\% \\
        BFD + KNN (K=15) & 0.00\% & 0.00\% & 0.00\% & 0.17\% \\
        BFD + K-CORN (K=15) & 0.00\% & 0.00\% & 0.00\% & 0.21\% \\
        BFD + KNN + K-CORN (K=15) & 0.00\% & 0.00\% & 0.00\% & 0.10\%
    \end{tabular}
\end{table}

Note that in Table \ref{tab:pilot-features}, all four datasets can be handled by multiple transportation modes. As a result, the values of features with an `` * '' are taken from their least restrictive mode. For example, the first dataset has two transportation modes. One of them can handle routes with up to four drops, and the other can only handle two-drop routes. So the least restrictive value (here, four) is reported for \textit{Max No. of Drops}.

Several other observations can be made from the results shown in Tables \ref{tab:pilot-features} and \ref{tab:pilot-results}, and Figures \ref{Figure2} and \ref{Figure3}:

\begin{itemize}
    \item More orders or transportation modes will increase the execution time.
    \item Generally speaking, the looser the constraints, the longer the execution time. So an increase in maximum number of drops, a larger average span of delivery time window, or a smaller difference between various truck capacities will increase execution time.
    \item The values of \textit{Max No. of Drops} show that all four datasets require 1PMD routes, but relying only on a 1P1D algorithm like BFD is not enough. So here, BFD is used to give us an upper bound for total cost and a lower bound for execution time.
    \item As we increase \textit{K} in KNN or K-CORN, we explore more of the solution space, and thus execution time increases. To diversify the search, it is better to use them together. 
    \item In the first three datasets, either KNN or K-CORN with ($K=10$) provide the optimal solution. In the fourth dataset and with gaps of $0.41\%$ and $0.52\%$ from the optimal value, respectively, they both satisfy the $5\%$ acceptable gap requirement. Nevertheless, we increased the value of \textit{K} and also considered their combination until reaching a $0.1\%$ gap from the optimal solution.
\end{itemize}

We did not specify how the values of $k$ in KNN or K-CORN are defined. Typically, CHR's planners control the intensity of the search by moving $k$ up or down in multiples of $5$. This iterative tweaking is informed by their knowledge of the problem's constraints and complexity.

Based on the results shown in Figures~\ref{Figure2} and \ref{Figure3} and Table~\ref{tab:pilot-results}, we use the combination of ``BFD, KNN ($K=15$), K-CORN ($K=15$)'', (henceforth known as BKK for simplicity) as the chosen algorithm for running other instances. The performance of BKK is then compared with existing software at CHR.

\subsection{Other Benchmark Examples}
First, three problem categories (easy, medium, and hard) are drawn from real CHR data based on \textit{total execution time}. We classify problems as ``Easy'' if we can solve them optimally in less than 600 seconds. ``Medium'' instances are those that cannot be solved using the Exact Method (even after one hour), but are solvable in less than 600 seconds by the BKK algorithm. Any problem where the total execution time of the BKK algorithm surpasses 600 seconds is categorized as ``hard''. Although the route generation step is generally the bottleneck, it's worth noting that the total execution time comprises route generation, SP model creation, and CPLEX solver run time.

Two different datasets in each of the three categories are considered. For each dataset, a summary of important features and execution time obtained by the BKK and Exact methods (if possible) are shown in Table \ref{tab:larger-features}. It is noteworthy that the Exact Method exceeds the run time limit of two hours for all the other datasets without obtaining any feasible solution.

\begin{table}
    % \small  % if we use this, the whole table will have small font
	\centering
	\caption{Features and results for larger test datasets}
	\label{tab:larger-features}
	\resizebox{\textwidth}{!}
	{% we can use one of the following:
    % \begin{tabularx}{\textwidth}{X|l|l|l|l|l|l}  % option 1
    % \begin{tabular}{p{0.3\linewidth}|l|l|l|l|l|l}  % option 2: text wrap
    \begin{tabular}{l|l|l|l|l|l|l}  % option 3: no text wrap and auto-resizing
         & \multicolumn{6}{c}{\textbf{Datasets}} \\ \cline{2-7} 
        \textbf{Features} & \textbf{1} & \textbf{2} & \textbf{3} & \textbf{4} & \textbf{5} & \textbf{6} \\ \hline
        No. of Orders & 336 & 1,279 & 961 & 590 & 3,129 & 5,365 \\
        Min Order Weight & 60 & 1.25 & 0.02 & 0.1 & 1 & 0.1 \\
        Avg Order Weight & 7,956 & 3,394 & 514.9 & 2,579.3 & 4,300 & 1,472 \\
        Max Order Weight & 19,991 & 14,980 & 16,850 & 47,999 & 39,997 & 40,000 \\
        Total Order Weight & 2,673,321 & 4,341,079 & 494,831 & 1,521,797 & 13,455,598 & 7,897,775 \\
        No. of Origins & 1 & 18 & 2 & 1 & 15 & 22 \\
        No. of Destinations & 28 & 538 & 451 & 342 & 1,274 & 2,369 \\
        Truck Capacity & 20,000 & 15,000 & 25,000 & 48,000 & 47,000 & 44,000 \\
        Max No. of Drops & 4 & 1 & 3 & 3 & 4 & 4 \\
        Max Distance (mile) & NA & NA & 3,000 & 5,000 & NA & 3,500 \\
        Max OOR Distance (mile) & 400 & NA & NA & 500 & 500 & 500 \\
        Max First to Last Drop Distance (mile) & NA & NA & NA & 1,000 & NA & 1,000 \\
        Avg Delivery Time Window Span (day) & 4.5 & 6 & 5 & 6.67 & 8 & 8 \\ \hline
        \textbf{Dataset Complexity} & easy & easy & medium & medium & hard & hard \\
        \textbf{Execution Time (sec) - Exact} & 1.5 & 0.8 & NA & NA & NA & NA \\
        \textbf{Execution Time (sec) - BKK} & 1.3 & 0.8 & 103.2 & 43.8 & 4,298.4 & 1,686.1
    \end{tabular}
    }
\end{table}

Comparing Tables \ref{tab:pilot-features} and \ref{tab:larger-features}, one can see minor modifications in the features. Because all six datasets only had one transportation mode (TL or LTL), there is only one \textit{truck capacity} and therefore no need to include \textit{No. of Transportation Modes}.
Also note that the unit for truck capacity and the order weight features of the datasets shown in Tables \ref{tab:pilot-features} and \ref{tab:larger-features} are the same (either \textit{pound} or \textit{kilogram}).

The initial results of BKK on all of these datasets were very promising. In fact, both KNN and K-CORN with ($K=10$) outperformed the existing technologies at CHR.

\section{Conclusions and Future Work}
We developed several solution strategies based on the SP formulation to solve many variants of rich VRP instances with different sizes and constraints for CHR. When tested on 10 initial datasets, the proposed algorithms outperformed the existing technologies at CHR.
Due to the success of this work, the proposed algorithms and framework, wrapped in a module called \textit{Optimizer}, are now fully integrated with CHR's Navisphere\textsuperscript{\textregistered} technology platform. 
CHR's managed services division (called TMC) is currently utilizing this technology for daily/weekly freight optimization across the transportation networks of large multinational shippers.

Optimizer gives users the flexibility to define all of the parameters and constraints of their problems, including route details and constraints related to types of vehicles, equipment, driver, product, location, and geographic area (among others). Each customer-specific optimization profile can be tailored to that customer's requirements.

It also provides them with an interface where they can select their desired algorithms (or a combination of them) from a list of available methods, as well as the intensity of the search by controlling their selected algorithms' parameters. As a side benefit of using Optimizer, the planners have considerably more time to focus on value-added tasks rather than manually building routes.
Optimizer is the de facto tool for transportation optimization at CHR. It is solving large complex problems, reducing costs, and delivering freight savings to C.H. Robinson's end customers.

Several interesting extensions for Optimizer capabilities are being studied. First, clustering-based approaches are being developed to deal with harder instances of the RVRP. Next, we are researching potential approaches to solve even more complex versions of CHR's problem (e.g., variations of the VRP with cross-docking). Developing column generation-based algorithms is another avenue of future research. Finally, we hope to explore ways to automatically select the best algorithm for a specific dataset given its size and features.

% Acknowledgments
\section*{Acknowledgment}
The authors are grateful to the C.H. Robinson's technology and commercial leadership for their support.
We also thank Steve Kravchenko and Michael Watson for their technical guidance which completely shaped the direction of the project.

\begin{appendices}
\section{Set Partitioning Model}
\label{sec:sp_model}
The SP model's notation and formulation is as follows:
\noindent\textbf{Sets}\\

$
\begin{array}{rl}
	I &\quad \mbox{set of orders}\\
	J &\quad \mbox{set of routes}\\
\end{array}
$\\

\noindent\textbf{Parameters}\\

$
\begin{aligned}
a_{ij} := \left\{
    \begin{array}{rl}
    1 &\mbox{ If order $i$ is covered by route $j$,} \\
    0 &\mbox{ Otherwise}
    \end{array} \right.
\end{aligned}\\
\begin{array}{rl}
c_j := \mbox{Associated cost of route $j$}\\
\end{array}
$\\

\noindent\textbf{Decision Variables}\\

$
\begin{aligned}
    x_j := \left\{
    \begin{array}{rl}
    1 &\mbox{ If route $j$ is selected,} \\
    0 &\mbox{ Otherwise}
    \end{array} \right.
\end{aligned}\\
$

\begin{align}
	\min & \sum_{j \in J}{c_j x_j} \label{eq:obj}\\
	\text{s.t:} \nonumber\\
	& \sum_{j \in J}{a_{ij} x_j} = 1 & \forall{i} \in I \label{eq:con1}\\
	& x_j \in \{0, 1\} & \forall{j} \in J \label{eq:con2}
\end{align}

The objective function minimizes the total transportation cost. Constraint~\eqref{eq:con1} ensures that each order is covered by exactly one route. Constraint~\eqref{eq:con2} states that all $x_j$ variables are binary. Note that in the simplest case, the cost of a route is calculated using the total route distance and the unit distance (mile or kilometer) cost. The unit distance cost depends on the origin and destination of a route. As a result and considering the network in Figure~\ref{Figure1}, there is no guarantee that the cost of a route A-B-C is less than the sum of the cost of route A-B and route A-C.
Therefore, we cannot use an equivalent set covering model where the equality in constraint~\eqref{eq:con1} is replaced with greater-than-or-equal-to.

\section{Pseudocode for Multi-drop Route Generation}
Algorithm~\ref{alg:md} gives the pseudocode for the multi-drop route generation procedure.

\begin{algorithm}[!ht]
	\caption{Pseudo-code for multi-drop route generation}
	\label{alg:md}
	\begin{algorithmic}
%	\linespread{0.75}\selectfont
		\State $K := $ maximum number of drops allowed; $ M := $ current route number of drops
		\State $ ODO := $ collection of origin-destination (OD) pairs with their respective orders \label{od-orders}
		\State $R, C$ = \textproc{1P1D Routes($ODO$)}
		\State $M = 2$
		\While{$M \le K$}
		    \State $R = R \,\ +  $ \textproc{1PMD Routes($R, C, M$)}
		    \State $M = M + 1$
		\EndWhile
		\State \Return $R$
		\Statex
		\Function{1P1D Routes}{$ODO$} \funclabel{alg:1p1d}
		\State $ R $ = empty list for keeping generated routes
		\State $ C $ = empty list for keeping generated order combinations
		\ForAll{$ p \in ODO $}
			\State $C = C \,\ +  $ \textproc{Order Consolidation($ p $)}
			\ForAll{$ c \in C $}
			    \State $R = R \,\ +  $ Generate and validate a route from $c$
		    \EndFor
		\EndFor
		\State \Return $R$ and $C$
		\EndFunction
		\Statex
		\Function{Order Consolidation}{$ p $}
		\State $ C $ = empty list for keeping generated order combinations
	    \State $ L := $ number of orders in $p$
	    \For{$l \gets 1$ to $L$}
	        \State $C = C \,\ +  $ Generate and validate every $l$-pair combination of orders in $p$
	    \EndFor
        \State \Return $C$
		\EndFunction
		\Statex
		\Function{1PMD Routes}{$R, C, M$} \funclabel{alg:1pmd}
% 		\State Use the feasible 1P(M-1)D routes to \textbf{generate feasible 1PMD routes}.
		\State $ FR = $ list of feasible 1P($M-1$)D routes obtained from $R$
		\ForAll{$ r \in FR $}
			\ForAll{$ c \in C $}
			    \State $R = R\,\ + $ Generate and validate a route by combining $r$ with $c$
		    \EndFor
		\EndFor
		\State \Return $R$
		\EndFunction
	\end{algorithmic}
\end{algorithm}

\end{appendices}

% \bibliography{rvrp}  %%% Remove comment to use the external .bib file (using bibtex).

% bibtex rvrpr


\end{document}